\documentclass[letterpaper, 10 pt, conference]{ieeeconf}  

\IEEEoverridecommandlockouts                              

\usepackage{graphics} 
\usepackage{amsmath} 
\usepackage{tcolorbox}
\usepackage{booktabs}
\usepackage{float}
\usepackage{cite}
\usepackage{caption}
\usepackage{hyperref}

\title{\LARGE \bf
Language-Guided Object Search in Agricultural Environments
}

\author{Advaith Balaji$^{1}$, Saket Pradhan$^{1}$, Dmitry Berenson$^{1}$
\thanks{This work was supported in part by the Office of Naval Research Grant N00014-24-1-2036 and NSF grants IIS-2113401 and IIS-2220876. $^{1}$Robotics Department, University of Michigan, Ann Arbor, MI, USA.
        {\tt\small [advaithb, saketp, dmitryb]@umich.edu}}%
}

\begin{document}

\maketitle
\thispagestyle{empty}
\pagestyle{empty}

\begin{abstract}
Creating robots that can assist in farms and gardens can help reduce the mental and physical workload experienced by farm workers. We tackle the problem of object search in a farm environment, providing a method that allows a robot to semantically reason about the location of an unseen target object among a set of previously seen objects in the environment using a Large Language Model (LLM). We leverage object-to-object semantic relationships to plan a path through the environment that will allow us to accurately and efficiently locate our target object while also reducing the overall distance traveled, without needing high-level room or area-level semantic relationships. During our evaluations, we found that our method outperformed a current state-of-the-art baseline and our ablations. Our offline testing yielded an average path efficiency of 84\%, reflecting how closely the predicted path aligns with the ideal path. Upon deploying our system on the Boston Dynamics Spot robot in a real-world farm environment, we found that our system had a success rate of 80\%, with a success weighted by path length of 0.67, which demonstrates a reasonable trade-off between task success and path efficiency under real-world conditions. The project website can be viewed at: \texttt{\href{https://adi-balaji.github.io/losae}{adi-balaji.github.io/losae}}

\end{abstract}


\section{INTRODUCTION}

Multiple studies investigating the stress experienced by farmers reported that the combination of the high workload and physical effort resulted in considerable levels of fatigue and mental strain \cite{ireland, strawberry}. Studies have also found that increasing mechanization and the use of automated machinery relatively reduces the workload experienced by farmers \cite{farm_mech}. Integrating robots into current farm environments can further reduce the workload experienced by farmers. If a robot can intelligently and efficiently search for an object or tool that a farmer needs, they will not have to expend any physical or mental effort looking for it themselves.

Several existing object search methods \cite{yokoyama2023vlfmvisionlanguagefrontiermaps, gadre2022cowspasturebaselinesbenchmarks, zhu2024navi2gazeleveragingfoundationmodels, chen2023traindragontrainingfreeembodied, Dorbala_2024, dharma, ziad}, leverage semantic information available to search through their environment, but the reason locating an unseen object in a farm environment is challenging and unique is that these farm environments are unstructured, and \textit{loosely} semantically organized. Homes and offices have a high degree of semantic organization offering rich semantic features from the floor and room level (\textit{e.g. } living room) down to the object level (\textit{e.g. } tv remote). Farms, gardens and similar outdoor environments are more loosely organized and lack the same degree of semantic information, preventing us from using richer semantic features that other semantic search methods use. However, it is reasonable to assume that objects are more likely to be located in proximity to other semantically related objects (\textit{e.g.} nail by a hammer) rather than alone (\textit{e.g} a single nail). This assumption allows using language to reason about an unseen object's likely location relative to seen objects.

\begin{figure}[t]
    \centering
    \includegraphics[width=0.475\textwidth]{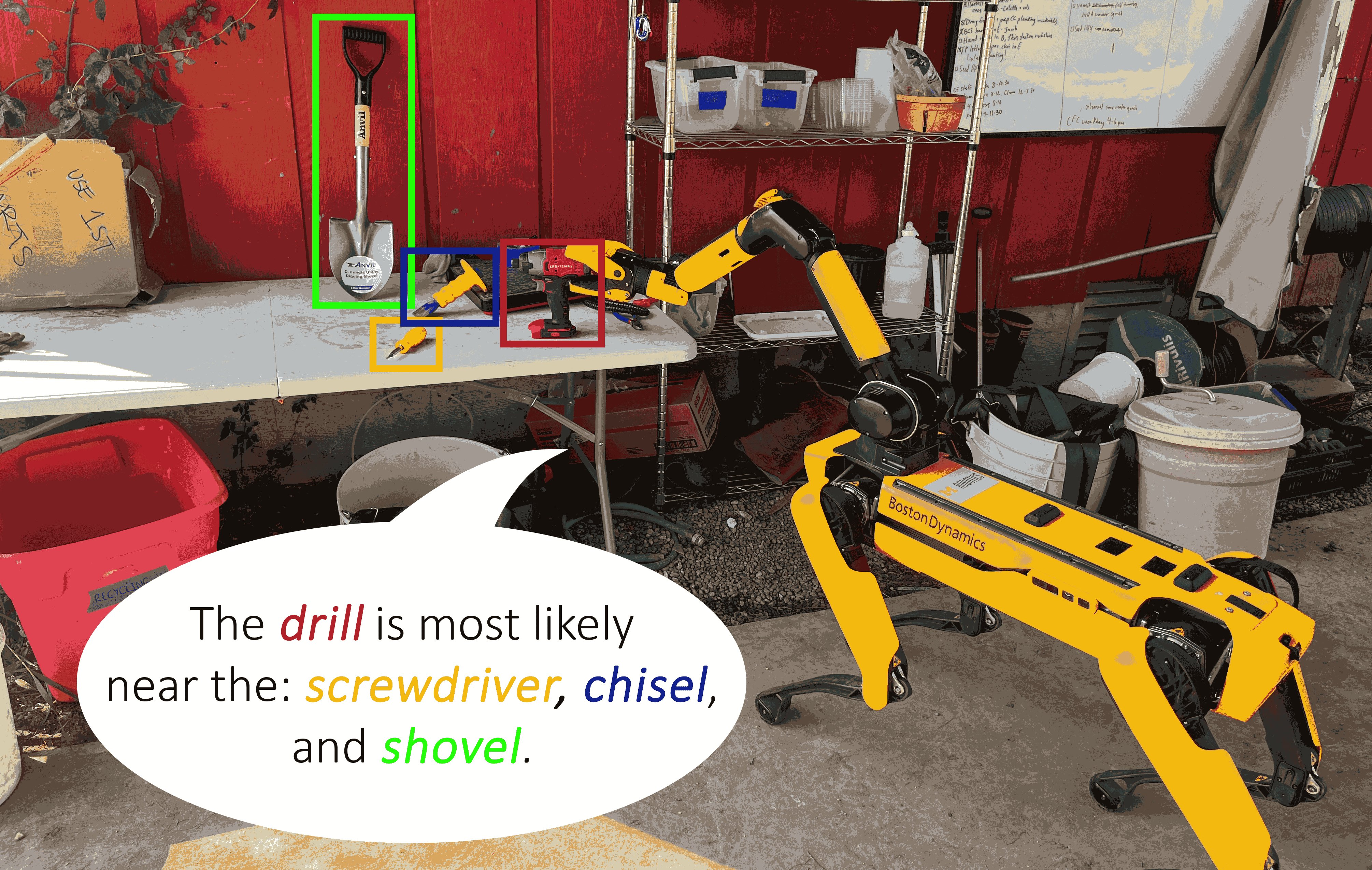}
    \caption{LOSAE allows the robot to find an unseen object using previously seen objects as instruments of reasoning. Here, the robot can understand that a tool like a \texttt{drill} is most likely located near similar tools like a \texttt{screwdriver}, \texttt{chisel} or \texttt{shovel}}
    \vspace{-5mm}
    \label{fig:attention_grabber}
\end{figure}

We present Language-Guided Object Search in Agricultural Environments (LOSAE), which uses only object-to-object semantic relationships inferred by an LLM to guide a robot for object search tasks in agricultural environments. The first step of LOSAE is for the robot to explore its environment and save the location of the objects it sees in its memory. When queried with an unseen target object, LOSAE uses an LLM to reason about which previously-seen objects are most likely to be near the target according to the semantic relationships between the seen object and target. LOSAE then plans a search path that allows the robot to accurately and efficiently find the target.  To our knowledge, this is the first method to address object search in loosely semantically organized environments, such as farms. We test LOSAE both using offline tests for the reasoning and path planning method and real-world tests using the Spot robot in a farm environment.

Our contributions are: 1) a novel method for object search in a loosely-semantically-organized environments; 2) a reasoning framework for object search leveraging only object-to-object semantic relationships from an LLM; and 3) experiments demonstrating the accuracy and efficiency of this method in a real farm setting.

\begin{figure*}[ht]
    \centering
    \includegraphics[width=\textwidth]{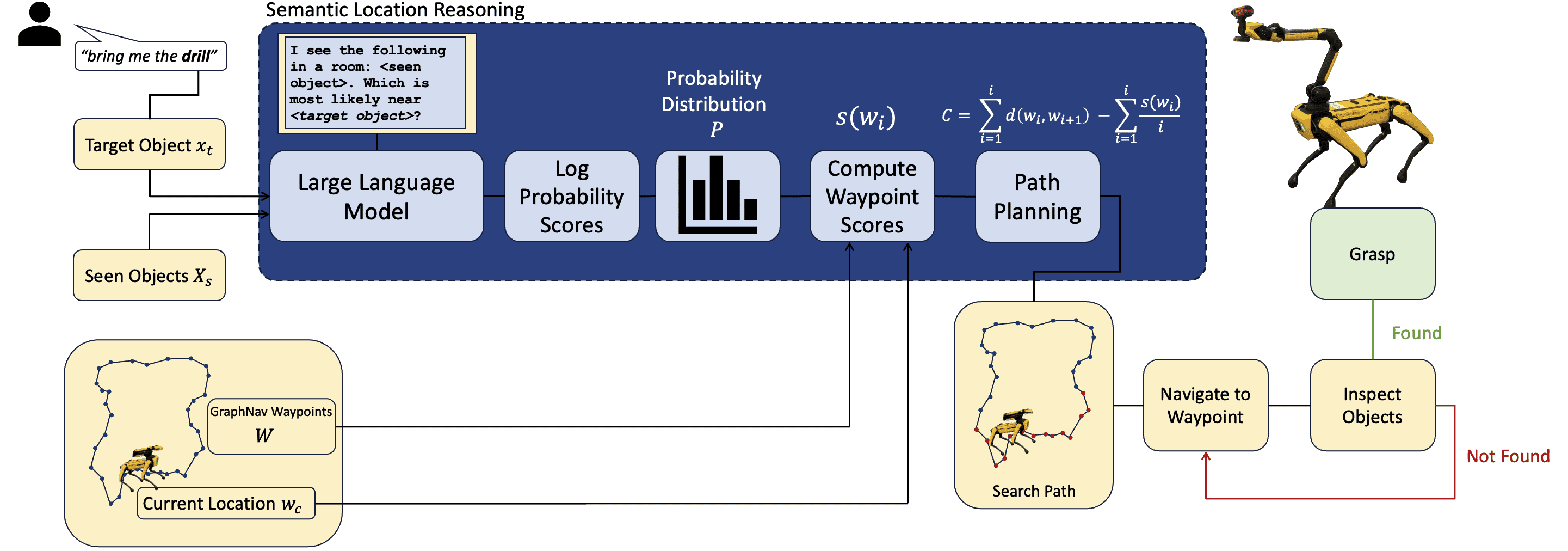}
    \vspace{-4mm}
    \caption{The robot is tasked with finding a target object $x_t$ based on a user query. The robot uses an LLM for semantic reasoning by generating a probability distribution $P$ based on the object-to-object relationships between a seen object $x_s$ in $X_s$ and the target object $x_t$. This distribution helps calculate waypoint scores $s(w_i)$ for each waypoint $w_i$. The robot plans a path through the waypoints according to the cost function $C$ that balances visiting high score waypoints and maintaining a short path length (see Methods for more details). The robot then navigates to a waypoint, inspects the objects around, and grasps the target if found; if not, it navigates to the next waypoint and continues the search.}
    \vspace{-3mm}
    \label{fig:system_diagram}
\end{figure*}


\section{RELATED WORK}

\subsection{LLMs for Robot Reasoning}
Recent advancements in Large Language Models (LLMs) have enhanced robot reasoning and decision-making \cite{llmgeneralservice}. LLMs have been used as semantic heuristics for navigation \cite{shah2023navigationlargelanguagemodels} and personalized robot assistance \cite{Wu_2023}, adapting to user-specific contexts. In object-goal navigation, an LLM-based zero-shot approach enables robots to search for objects without prior training \cite{Dorbala_2024}. Our work builds on this, focusing on the unique challenges of agricultural environments with looser semantic organization.

\subsection{Semantic Representations for Object Search}
Semantic understanding has advanced object search capabilities, with methods like goal-oriented semantic exploration and language-driven actions \cite{Chaplot,gadre2022cowspasturebaselinesbenchmarks}. LLMs and Vision-Language Models (VLMs) can also be used to gauge semantic affinities between objects for mechanical search of occluded objects, a method we exploit for our problem \cite{sharma2023semanticmechanicalsearchlarge, chen2023traindragontrainingfreeembodied}. Approaches described in \cite{rosinol, ginting2024seeksemanticreasoningobject, ginting2024semanticbeliefbehaviorgraph, zhu2024navi2gazeleveragingfoundationmodels, jatavallabhula2023conceptfusionopensetmultimodal3d} leverage 3D scene graphs and high-level semantic cues for navigation, but these rely on hierarchical relationships less common in agriculture. We aim to enhance semantic object search in farms and gardens by combining LLM-based reasoning with object-to-object relationships.

\subsection{Robotics in Agriculture}
The application of robotics in agriculture has seen significant growth. Existing work in agricultural robotics focuses on tasks such as crop monitoring, harvesting, and soil analysis \cite{Freeman-2023-135014, Freeman-2023-137532}. \cite{sivakumar2024lessonsdeployingcropfollowundercanopy} Focuses on using semantic ``keypoints'' to navigate under corn canopies, but does not extend to broader semantic navigation. \cite{hutter, gehring} explored generic solutions and using quadrupeds for industrial site inspection, which shares similarities with agricultural inspection tasks. Our work extends this domain by addressing the specific challenge of object search in farms.

\section{PROBLEM STATEMENT}
Our objective is to allow a robot to navigate to and locate an unseen target object $x_t$ (an object that was not previously detected and localized by the robot) identified by a text label $l_t$ in an agricultural environment $E$ containing a set of seen objects $x_s \in X_s$, each described by text labels $l_s \in L_s$ and associated locations. We assume access to a Large Language Model capable of producing log probability scores for each output token, and we assume our environment is \textit{loosely} semantically organized, which we define as follows. 

In highly semantically organized environments such as homes or offices, several methods represent the semantic scenes as a hierarchy (or a hierarchical graph) of high-level entities to low-level entities \cite{li2024pixelsgraphsopenvocabularyscene, gu2023conceptgraphsopenvocabulary3dscene, werby2024hierarchicalopenvocabulary3dscene}. If $H$ denotes the set of all semantic labels available for this scene, then there exists a mapping function $f: H \rightarrow \{H_1, H_2,\dots, H_n\}$ where $H_n$ is the set of semantic labels at the $n^{th}$ hierarchical level. In a \textit{loosely} semantically organized scene, $H_n$ for $n>1$ starts to become sparse or undefined. This means we assume our system only has access to an $H_1$ level of semantic scene understanding. This is reasonable to assume as environments like farms and gardens have distinct and unique lower-level objects and tools but less distinct larger areas or rooms. Therefore, we assume that the probability of locating $x_t$ at any given $x_s$, represented as $p(x_t \mid x_s)$, depends mainly on lower-level, object-to-object semantic relationships. We also assume $x_t$ exists at only one $x_s$ at any given time.  

Finally, we assume that the robot has access to an egocentric RGB-D camera and a map $G=(W,e)$ of the environment, that it can navigate through autonomously, structured as a graph of waypoints (or nodes) $W$ and edges $e$. We also assume the robot has a manipulator arm that is capable of grasping an object. We consider $x_t$ to be successfully located if the robot can come within a small radius $\epsilon$ of the object. Evaluating methods that address this problem include calculating Success Rate (SR), Path Efficiency (PE), and Success weighted by Path Length (SPL) \cite{on_eval_of_emb_nav}.

\section{METHODS}
LOSAE consists of an Environment Exploration phase, a Location Reasoning phase, and an Object Search phase. As illustrated in Figure \ref{fig:system_diagram}, after first recording $L_s$ for all seen objects $X_s$ in the Exploration Phase, the robot is tasked with finding $x_t$. In the Location Reasoning phase, it will construct a discrete probability distribution $P = p(x_t, X_s)$ according to an LLM's output, representing the likelihood of $x_t$ being located at each $x_s$. Using these likelihoods, it assigns a score $s(w_i)$ for each waypoint $w_i$. During the Object Search phase, the robot plans a path through the waypoints according to a cost function $C$ that balances path length and visiting high-probability waypoints early. If $x_t$ is successfully found in $w_i$ near a particular $x_s$, the robot will attempt to grasp $x_t$ and bring it back to its start location. The object is considered lost if it is not found after consuming 95\% of the probability while searching along the path.

\subsection{Environment Exploration} 
In order to reason about the location of an unseen object, the robot must first populate $L_s$. This will be used as input along with $l_t$ to construct our probability distribution that helps plan the path to search for $x_t$. To do this, we produce object detections in real-time while traversing our environment once. We use a pre-trained YOLOv8 object detection model capable of producing bounding boxes with semantic labels for the various tools and objects we are working with \cite{yolov7trainablebagoffreebiessets, yolov8}. We opted to use a pre-trained YOLO model as opposed to an open-vocabulary detector like GroundingDINO \cite{groundingdino} or OWL-ViT \cite{owlvit} as we found that these models tend to perform poorly in unorganized environments such as the one we operate in. This performance drop is further amplified when dealing with uncommon objects such as farm tools. However, our system can still be used with an open-vocabulary exploration phase by attempting to semantically label observed segmentation masks using Segment-Anything and VLMs \cite{sam, chen2023palijointlyscaledmultilinguallanguageimage, sharma2023semanticmechanicalsearchlarge}. While traversing the environment, we sweep the camera from left to right while passing each RGB image through the the object detector. We register any unique $l_s$ to $L_s$, and save the 3D location of $x_s$ relative to the robot’s body frame.

\subsection{Location Reasoning}
The core of our approach involves reasoning about which $x_s$ are most likely to be in proximity to the unseen $x_t$. Similar to how people will attempt to search for an object in areas that are occupied by other related objects to have a higher chance of finding their target object, we aim to allow the robot to semantically understand which $x_s$ to search near to most likely locate $x_t$. To achieve this, we make use of an LLM to compute a probability distribution $p(x_t \mid X_s) = p(x_t \mid x_s), \forall x_s \in X_s$ representing the probability of $x_t$ being located at each individual $x_s \in X_s$, drawing from a similar semantic reasoning method \cite{sharma2023semanticmechanicalsearchlarge}. 

To calculate a single $p(x_t \mid x_s)$, we first prompt an LLM with the instructions outlined as follows along with $l_t$ and $l_s$ as input. We use the OpenAI GPT-4o mini model with a 0 temperature setting for our system. 

\begin{tcolorbox}[colback=blue!5!white, colframe=blue!75!black, title=System Prompt]
You are an expert object location reasoning robot. You will be given some seen objects and a target object. You need to output which is the best seen object to go to in order to find the target object. You may only use the seen objects for reasoning, and must output a seen object to go to.
\end{tcolorbox}
\begin{tcolorbox}[colback=blue!5!white, colframe=blue!75!black, title=User Prompt]
I see the following: \textless $l_s$\textgreater. Where should I go to find \textless $l_t$\textgreater?
\end{tcolorbox}

 The LLM outputs an answer to the prompt along with a set of log probability scores for each output token in the response. A log probability score $\log p(t_n \mid t_0, t_1,\dots, t_{n-1})$ represents how probable it is that a token $t_n$ appears after a prior sequence of tokens $(t_0, t_1,\dots, t_{n-1})$ according to the LLM. These scores come from the model's final layer, where the hidden state $h_n$ passes through a linear layer, producing a logits vector $z_n$. Softmax converts $z_n$ into a probability distribution over the vocabulary, and the highest-probability token is selected as the output token. The natural log of its probability gives us the aforementioned log probability score. Upon collecting each output token's score, we calculate $\log p(x_t \mid x_s)$ to be the average log probability across the output tokens. We then exponentiate to obtain  $p(x_t \mid x_s)$: 

 \vspace{-5mm}
 $$p(x_t \mid x_s) = \exp(\frac{1}{N} \sum_{n=0}^{N} \log p(t_n \mid t_0, t_1,\dots, t_{n-1})) \eqno{(1)}$$

\noindent Limiting the input to just one $l_s$ at a time allows us to extract $p(x_t \mid x_s)$ for all $x_s$ independently \cite{sharma2023semanticmechanicalsearchlarge}. In contrast to \cite{sharma2023semanticmechanicalsearchlarge}, we found that in some cases, depending on the prompt and model used, the exact token for $x_t$ may not exist, or exists and is null. Therefore considering the average token probability of the entire output sequence presents a more reliable scoring mechanism to represent $\log p(x_t \mid x_s)$. We compute and collect $p(x_t \mid x_s)$ for all $x_s \in X_s$ and normalize to finally obtain our complete probability distribution $p(x_t \mid X_s)$. This distribution can be used to score waypoints based on their semantic affinity to the $x_t$, which will help plan a path to accurately locate $x_t$.

\subsection{Object Search}

To complete our task of finding $x_t$, we can use the constructed probability distribution $p(x_t\mid X_s)$ to plan a path through the environment that will most successfully and efficiently find $x_t$. We aim to visit locations in the environment that have a high probability of containing $x_t$ while still traversing a short overall distance. 

First, we establish the method to calculate the probability of $x_t$ being near some spatial location $w$ described by Cartesian coordinates $(w_x, w_y)$ relative to the robot's start location. Each object $x_s$ in the environment has an associated probability of finding $x_t$ there, and each object exists at or near some spatial location $w$ in the environment. Therefore, the probability of finding $x_t$ at any given spatial location $w$ depends on the number of objects in that location as well as the probability score for each object there. Since the robot is limited to navigating to specific waypoints (or nodes) in its graph map $G$, and we assume that the events of $x_t$ being at $x_s, \forall x_s \in X_s$ are mutually exclusive, we define the probability of finding $x_t$ at a given location, or waypoint $w_i$, to be 

\vspace{-5mm}
$$s(w_i) = \sum_{i=0}^{N}  p(x_t | x_{s_i}) \eqno{(2)}$$
\vspace{-2mm}

\noindent where $N$ is the number of seen objects $x_s$ located at waypoint $w_i$. This represents the sum of the individual probability scores for each object at $w_i$. Upon calculating the waypoint score $s(w_i)$ for all waypoints, we can plan an efficient path through the environment that reduces the distance traveled while also prioritizing searching near high-probability areas. This can be achieved by planning a path according to the cost function as follows.

$$C(W_p) = \sum_{i=1}^{N-1} d(w_i, w_{i+1}) - \sum_{i=1}^{N} \frac{s(w_i)}{i} \eqno{(3)}$$

\noindent The cost $C$ of a path $W_p$ represented as a sequence of waypoints $w$ is the path length (according to the geodetic distance function $d$) minus a contribution from the waypoint scores, where higher scores are desirable early in the path. This cost ensures the robot selects a path that trades off visiting high-probability waypoints early and shortening the overall travel distance, balancing both efficiency and accuracy.

In our environment, since the number of waypoints containing objects are relatively small, we choose to find every possible permutation of a path, compute its score according to $C$, and select the path that has the minimum cost. This method guarantees the optimality of our path according to the cost function, but scales by $n!$ and would not be feasible for larger maps. If presented with a larger map, one could use a version of the A* planning algorithm to find a path using our cost.

Once the robot has the path, we navigate the robot directly to the first waypoint to examine $w_e$ and set the camera view to the previously saved pose of any object $x_i$ seen at $w_e$. We compute a bounding box centroid for $x_s$ using our object detection model and then navigate the robot to move within a threshold distance of the object.

Once the robot reaches the vicinity of $x_i$, we use the camera to scan for $x_t$ (see Results for more details). While scanning, object detections for $x_t$ are computed in real-time, and are filtered out if their confidence is less than 0.8. If there were no bounding boxes detected that were confident enough, we assume $x_t$ is not located at $x_i$. After scanning near all objects at $w_e$, if there are no confident bounding boxes computed for $x_t$, we assume $x_t$ does not exist at $w_e$, go to the next waypoint in the path, and repeat the search process. If we do detect a confident enough bounding box for $x_t$, we compute its centroid and move towards it. At this stage, if the robot is truly within a radius $\epsilon$ to $x_t$, the object is considered successfully located. 

We also include an additional step of grasping the object and bringing it back to the start location. We compute the segmentation mask of $x_t$ by prompting Fast-SAM with the bounding box, and segment the depth image according to the mask output \cite{sam, fast-sam}. Using the camera intrinsics, we re-project the depth segment into world space and transform it to the robot's body frame, producing a point cloud of $x_t$. This point cloud is then passed to the Grasp Pose Detection model \cite{gpd1, gpd2} which produces a 6-DOF grasp pose that becomes the target for grasping. Once grasped, the robot brings the object back to the initial waypoint.

\section{RESULTS}

To judge the system's ability to accurately and efficiently find an unseen object among seen objects in an agricultural environment, we conducted qualitative and quantitative experiments by deploying our system on a real robot and running it in a real-world farm environment. We performed these experiments against a method similar to a state-of-the-art (SOTA) baseline. The robot we used for experimentation was the Boston Dynamics (BD) Spot with the Spot Arm payload. All models used during any stage of the pipeline were run on a Lenovo Legion 5 laptop equipped with an NVIDIA GeForce RTX 3050 GPU, except for the GPT-4o mini model that was called through OpenAI's API. We also performed some tests offline to evaluate our core reasoning method against the same baseline and 2 ablative methods.

\subsection{Our Environment} 
\begin{figure}[t!]
    \centering
    \includegraphics[width=0.475\textwidth]{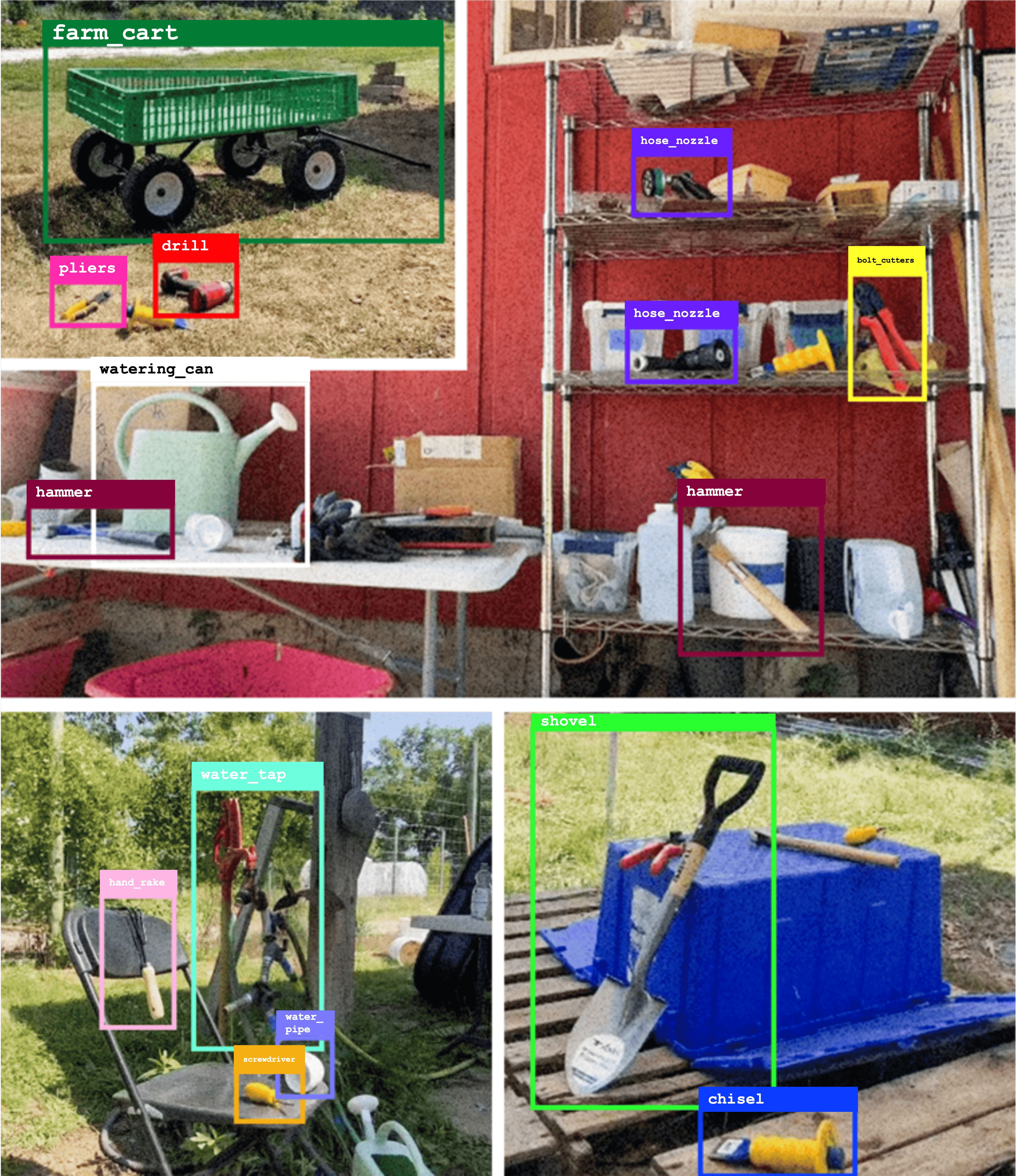}
    \vspace{1mm}
    \caption{Images from our custom YOLOv8 dataset showcasing the objects and environment we work with. Top left: \texttt{farm cart}, \texttt{drill}, and \texttt{pliers} by the field. Middle: \texttt{watering can}, \texttt{hammer}, \texttt{water hose nozzle}, and \texttt{bolt cutters} by the tool shelf and table in the tool storage. Bottom left: \texttt{water tap}, \texttt{water pipe}, \texttt{hand rake} and \texttt{screwdriver} by the water station. Bottom right: \texttt{shovel} and \texttt{chisel}.}
    \vspace{-5mm}
    \label{fig:all_scenes}
\end{figure}

We deployed the robot in an agricultural environment at the University of Michigan Campus Farm. Some pictures of our environment and objects can be seen in Figure \ref{fig:all_scenes}. The area consisted of a ``\texttt{tool storage}'' and work area with shelves and tables, a ``\texttt{water station}'' with a hose, a well pump and water tap, a ``\texttt{wash station}'' next to the water station with tubs and tables for washing harvest, and a ``\texttt{harvest station}'' by the field with bins for vegetables and a farm cart for transport. The entire area was approximately 900 $\text{m}^2$. A typical map generated here consisted of 20-40 waypoints. As referenced in the problem statement, in this environment, the robot is only able to perceive lower, object-level labels belonging to $H_1$ (such as ``\texttt{hammer}''). Semantic labels $H_n$ for $n>1$ (such as an area ``\texttt{tool table}'') do not exist in our environment as the environment has no distinct boundaries between areas or rooms like the \texttt{tool storage}, \texttt{wash station}, etc. Even if there was, any object $x$ does not always have a strong relationship with its higher-level semantic label due to the high movability of these objects. 

\subsection{Metrics} 
We report results for our real-world experiments and core reasoning method independently. The metrics we chose for the real-world experiments are Success Rate (SR) and Success weighted by Path length (SPL), and report each metric per full episode. An episode is defined as one full object locating sequence from query to successful location. The Success Rate is calculated as the number of successes divided by the number of attempts, where a success is defined as the robot arriving within $\epsilon=0.5$ m of the target object, which is sufficient to execute the grasping policy. The SPL offers a method to score the success and the efficiency of the robot’s path by comparing it to the shortest path to the target object and is calculated as 

\vspace{-2mm}
$$SPL = \frac{1}{N} \sum_{i=1}^{N} S_i \frac{p_s}{\max(p_i, l)},$$
\vspace{-2mm}

\noindent where $N$ is the total number of episodes or attempts, $S_i$ is a binary indicator that is 1 for success and 0 if not, $p_s$ is the length of the shortest possible path from the start location to the target object, and $p_i$ is the actual length of the path traversed by the robot to reach the goal \cite{on_eval_of_emb_nav}. To evaluate the reasoning method, we evaluate the Path Efficiency (PE). This is calculated as $PE = \frac{p_s}{p_i}$, where $p_s$ is the ideal shortest path length from the start to the goal, and $p_i$ is the length of the predicted path from the start to the goal.

\subsection{Baseline} A different solution to a problem very similar to ours was explored in \cite{ginting2024seeksemanticreasoningobject}. Their method, SEEK, leverages a known environment to construct a room-level semantic search policy that uses the room names to reason about an unseen object's location. The room reasoning phase consists of building a Relational Semantic Network (RSN) off of LLM outputs that is capable of relating a room label to a probability of the target object existing there and computing a global robot policy that navigates the robot from its current location to the most probable room location (while balancing distance traveled, similar to our method). An active semantic search policy allows the robot to find the target object in the room once reached by executing a policy that minimize uncertainty of the target's location \cite{ginting2024semanticbeliefbehaviorgraph}. We sought to use SEEK as a baseline, however, the code was not available and it was not feasible to implement their RSN and active semantic search from scratch. We did not have access to a reliable test environment with its blueprint to faithfully recreate its exact global room search policy. Additionally, the entropy-based semantic search method relied on accurately localizing an object's pose using the Spot LIDAR, which we did not have access to either. Thus we created a baseline that was designed to be similar to SEEK, which we call Room Search. 

\textbf{Room Search}: First, we establish some rough ``rooms'' in our environment. As mentioned earlier, the divisions between rooms can become blurry in spaces such as farms and gardens; however, it is reasonable to assume that there exist some rough areas that can be grounded as semantic rooms. We define the ``rooms'' to be  
\texttt{tool storage}, \texttt{wash area}, \texttt{water station}, and \texttt{harvest station}.

SEEK pre-trains the RSN from LLM output, as their method was meant to work in environments without internet and large computational resources. Since we have access to the internet and the OpenAI API, we can instead make use of the ChatGPT4 model directly by querying it to output a probability score for each room according to the target object, which is the way SEEK generates the dataset for their RSN. We also set up an ``active semantic search controller'' that aims to search areas of high semantic utility while updating its belief of the target object location iteratively. 

The robot is given the waypoints of the semantic rooms and initialized at a random waypoint. It will use the LLM to produce a probability score for each room and plan a path that trades off the distance traveled and probability consumed. Once it reaches a room, the robot will collect bounding box labels for everything it sees in its field of view and compute the semantic similarity (cosine distance between word embeddings) between the target object label and bounding box labels using a Language Embedding Model \cite{clippaper, bert, cosdist}. We use DistilBERT for its quick computation time \cite{distilbert}. Using these scores, the robot will search for the target object in an area of high semantic similarity. If the target object is not found there, the robot will update its belief about the goal location, and search a new area. If the robot exhausts areas to search, it will move to the next room.  

\textbf{LOSAE:} The robot first executes the Environment Exploration. Due to the limited range of the Spot gripper depth camera, instead of saving the relative position of the object, we save the waypoint and the arm pose at which the object was seen at, so that we can later return to the same waypoint and arm pose and walk to the object using the BD API \cite{bostondynamics2023spot}. After populating the object memory, the robot is placed at a random waypoint and then queried with an unseen object. The robot will reason about the location, plan a path, and traverse the path visiting various locations as it walks. When it visits an object at a waypoint, the arm is deployed to scan around the object by sweeping a 120-degree angle from a top-down view. If the object is found, we consider the episode a success and record the robot's traversed path from its initial location to its final location. We also attempt to grasp the object and bring it back to the initial location, which is done using the Grasp Pose Detection (GPD) model (as explained in the Methods section), or the BD API if GPD fails \cite{pas2017graspposedetectionpoint}. We do not consider grasp failures as part of our success metric as grasping an object reliably in an unstructured, cluttered environment like ours constitutes a different, challenging problem. We choose to focus on semantic reasoning and object search.

We ran each method 15 times, initializing the location of the target object and the robot to within 2 m of a random waypoint each run. The metrics are reported in Table \ref{table1}.

\begin{table}[h]
\centering
\setlength{\tabcolsep}{21pt} 
\begin{tabular}{lcc}
\toprule
\textbf{Methods} & \textbf{SR $\uparrow$} & \textbf{SPL $\uparrow$} \\
\midrule
\textbf{LOSAE (Ours)}           & \textbf{0.80} & \textbf{0.67} \\
Room Search                       & 0.73          & 0.51 \\
\bottomrule
\end{tabular}
\caption{SR and SPL of LOSAE compared to Room Search tested on 15 trials per method}
\vspace{-2mm}
\label{table1}
\end{table}

\begin{figure}[t]
    \centering
    \includegraphics[width=0.475\textwidth]{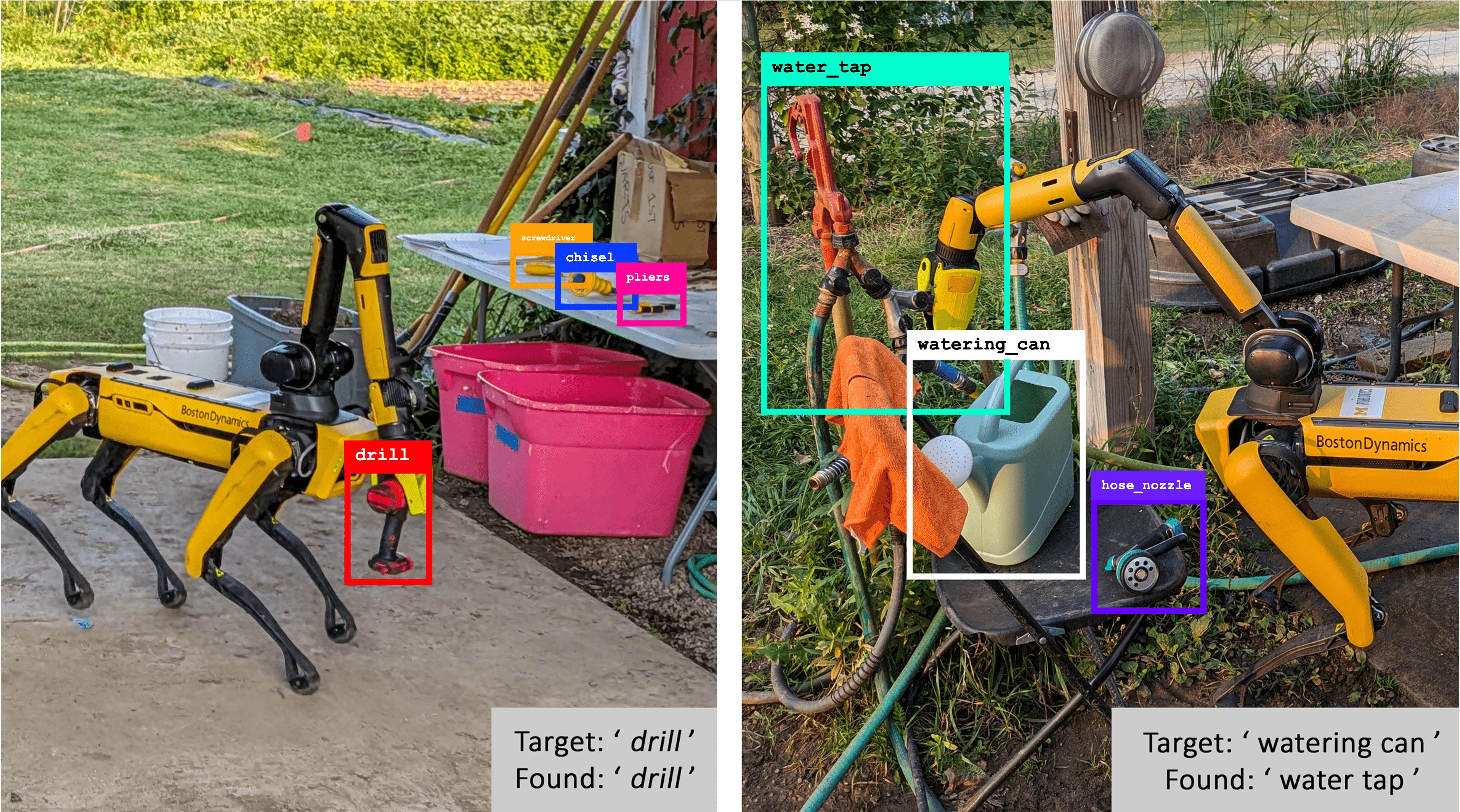}
    \caption{Left: an instance of the robot correctly identifying the target object at the correct location. Right: an instance of a false positive. The robot correctly finds the watering can next to the water tap and water hose nozzle, but grasps the water tap due to perceptual errors. }
    \vspace{-6mm}
    \label{fig:tf_comp}
\end{figure}

The data shown in Table \ref{table1} demonstrates LOSAE's strong capabilities of accurately locating an unseen target object in a real farm environment. We outperform the Room Search method, reporting an SR of $80\%$ and SPL of $0.67$. LOSAE is able to overcome the need for higher level semantic descriptions and utilize only object-level labels to accurately find an unseen target object. Failures often stem from false positives—where YOLO incorrectly classifies an object, leading the robot to an incorrect search location—and false negatives, where YOLO fails to detect the target, preventing a successful search altogether. Figure \ref{fig:tf_comp} illustrates an example of a false positive that affected performance. Given that objects are always moving and changing in a farm, real world performance of the perception system could be hindered by issues such as occlusions, lighting, and changes in object orientation.
 
Therefore, we independently compare our core reasoning method against Room Search and 2 ablative methods. The Hottest Object method simply chooses the highest probability object and searches around that object. The Hottest Waypoint always chooses the waypoint with the highest probability sum and searches for the target object around that waypoint. This allows us to understand the performance independent of a robot's capabilities. We ran the reasoning and path planner for each method offline 15 times, initializing the location of the target object and robot to a random waypoint each run.

 \begin{table}[h]
\centering
\setlength{\tabcolsep}{21pt} 
\begin{tabular}{lcc}
\toprule
\textbf{Methods} & \textbf{Avg. PE (Std. Dev) $\uparrow$} \\
\midrule
\textbf{LOSAE (Ours)}           & \textbf{0.84 (0.27)} \\
Room Search                       & 0.72 (0.33) \\
\midrule
Hottest Object                       & 0.21 (0.32) \\
Hottest Waypoint                       & 0.19 (0.41) \\
\bottomrule
\end{tabular}
\caption{PE of LOSAE compared to Room Search and ablative methods tested on 15 trials per method}
\label{table2}
\end{table}

The data shown in Table \ref{table2} reports the Average PE and the Standard Deviation for each method. LOSAE produces a path that is on average 84\% as efficient as the ideal path. It is possible that Room Search fails to attain the same level of path efficiency because this method relies on room probability scores directly generated by an LLM. It could mistakenly set the probability too high for some room, encouraging the path planner to travel long distances. Gauging the LLM's confidence in its answer (see Location Reasoning) may provide a more reliable means of scoring semantic relationships. Additionally, the PE for Room Search and could could reduce even further when the active semantic search procedure is taken into consderation. We also find that always choosing the highest probability object or waypoint regardless of distance greatly reduces PE. This is expected as the robot would disregard the distance cost completely, resulting in longer paths. Considering both the waypoint score and distance cost while planning a path offers a good PE while still outperforming Room Search.

\section{CONCLUSION AND FUTURE WORK}

This paper presents a novel method to allow a robot to find an unseen target object among a set of seen objects in a farm environment that is loosely semantically organized. We eliminate the need for higher-level semantic labels and leverage object-to-object semantic relationships inferred by an LLM to guide a path planner to efficiently search the environment to successfully find the target object. We evaluated the performance of our method by deploying it in a real-world agricultural environment on the Boston Dynamics Spot platform, finding that LOSAE achieves a success rate of 80\% and an SPL of 0.67 for our problem setup. We also evaluated the performance of the core reasoning method offline and found that LOSAE produces a path that is on average 84\% efficient relative to the ideal shortest path.

Our method relies solely on language for reasoning. Future work could integrate visual features and VLMs to enable deeper reasoning for object search. Additionally, our reasoning occurs only at query time; incorporating real-time observations with LLMs in a closed-loop manner could improve search efficiency and adaptability.




\bibliographystyle{IEEEtran}
\bibliography{biblio}

\addtolength{\textheight}{-12cm}   

\end{document}